\documentclass{article} % For LaTeX2e
\usepackage{iclr2027_conference,times}

\usepackage{amsmath,amsfonts,bm}

\def\eqref#1{equation~\ref{#1}}
\def\1{\bm{1}}

\def\vx{{\bm{x}}}

\DeclareMathAlphabet{\mathsfit}{\encodingdefault}{\sfdefault}{m}{sl}
\SetMathAlphabet{\mathsfit}{bold}{\encodingdefault}{\sfdefault}{bx}{n}

\usepackage{enumitem}
\usepackage{hyperref}
\usepackage{url}
\usepackage{graphicx}
\usepackage{subcaption}
\usepackage{booktabs}
\usepackage{float}

\usepackage{xcolor}

\newcommand{\syn}{\mathrm{syn}}
\newcommand{\real}{\mathrm{real}}
\usepackage{amsmath}
\usepackage{amssymb}
\usepackage{amsthm}

\usepackage{centernot}

\newtheorem{proposition}{Proposition}[section]

\title{Representational Change Under Self-Consuming Training: A Sparse Autoencoder Analysis}

\author{Suchit Gupte, Zhongteng Cai, Xueru Zhang, Mohammad Mahdi Khalili 
\\
Department of Computer Science\\
The Ohio State University\\
Columbus, OH 43210, USA \\
\texttt{\{gupte.31, cai.1125, zhang.12807, khalili.17\}@osu.edu} \\
}

\iclrfinalcopy % Uncomment for camera-ready version, but NOT for submission.
\begin{document}

\maketitle

\begin{abstract}
Large language models are increasingly exposed to model-generated text, raising the possibility that future models will be trained recursively on the outputs of their predecessors. Prior work on such self-consuming training loops has largely focused on output quality and downstream performance, leaving their effects on internal representations poorly understood. We study how representations evolve across five generations of self-consumption and whether sparse autoencoders (SAEs) remain reliable under this shift. At each generation, we compare a lineage trained on synthetic data with a matched control trained on real data under the same training settings. We evaluate how well SAEs trained before self-consumption transfer to later generations, measuring reconstruction and sparsity across SAE widths, architectures, and model layers. We then fine-tune the original SAE on each generation and use sparse probing to test whether task-relevant information remains concentrated in a small number of latents or becomes more distributed across the representation. We find that zero-shot SAE performance degrades progressively under self-consumption, with growing reconstruction and sparsity gaps that vary substantially across layers. Fine-tuning largely restores reconstruction, but not feature-level usefulness: task-relevant information remains accessible in the full latent space while becoming less concentrated in individual features. These results show that self-consumption induces representational drift and that reconstruction fidelity alone is insufficient to establish that an adapted SAE remains equally useful for interpretation.
\end{abstract}
\section{Introduction} 
Language models are increasingly trained on corpora that contain text generated by other language models. As high-quality human-written data becomes harder to scale~\citep{villalobos2024will}, machine-generated text is accumulating rapidly on the web and even in scientific writing~\citep{liang2024mapping}. Future models are therefore likely to be trained, at least in part, on the outputs of their predecessors. Repeating this process creates a \textit{self-consuming loop}~\citep{alemohammad2024self}, in which one model generates training data for the next, and the cycle continues across generations. Prior works show that such recursive training can progressively narrow the learned distribution and erode its tails, a phenomenon known as model collapse~\citep{shumailov2023curse,shumailov2024ai}. However, these effects have primarily been studied through externally observable quantities such as likelihood, output diversity, and downstream performance (Refer Sec~\ref{sec:related}).

This leaves a complementary question largely unexplored: what happens to the model’s internal representations under self-consumption? This question is particularly important for interpretability, since many methods operate directly on model internals~\citep{alain2016understanding,belinkov2022probing,sharkey2025open}. Sparse autoencoders (SAEs) have become a widely used tool for decomposing dense activations into sparse, individually interpretable features~\citep{bricken2023monosemanticity,cunningham2024sparse,templeton2024scaling,gao2025scaling}. As models evolve through repeated training on synthetic data, however, it is unclear whether an SAE trained on an earlier model will continue to transfer, or whether the features recovered after adaptation retain the same usefulness. We therefore use SAEs as a diagnostic of representational change and ask two questions:

\begin{enumerate}[leftmargin=3em,rightmargin=3em,itemsep=1pt,topsep=2pt]
    \item \textbf{Does the original SAE still transfer?} SAEs often transfer between related models, such as base and chat models~\citep{kissane2024saes,lan2024sparse, gupte2026pruningmeetsinterpretabilitypreserving}, but whether they survive a self-consuming loop is unknown. Progressive failure of a fixed SAE is evidence of representational drift.
    \item \textbf{Does adapting the SAE restore its usefulness?} Fine-tuning the original SAE on each new generation may restore reconstruction, but good reconstruction does not guarantee that information remains localized in individually useful features.
\end{enumerate}

We study these questions over five generations of self-consumption training using \texttt{Pythia-160M}~\citep{biderman2023pythia} and \texttt{Gemma-2-2B}~\citep{gemmateam2024gemma}. For \texttt{Pythia-160M}, we evaluate multiple SAE widths and architectures; for \texttt{Gemma-2-2B}, we examine early, middle, and late layers. Each synthetic lineage is paired with a real-data control trained with the same token budget, same number of optimization steps, and same number of generations, isolating the effect of training on synthetic rather than real data. 

Across these matched lineages, we observe a consistent progression. Zero-shot SAE transfer degrades under self-consumption: on \texttt{Pythia-160M}, synthetic--real gaps in reconstruction and sparsity grow across generations for all SAE widths and architectures, while on \texttt{Gemma-2-2B} the effect is strongly depth-dependent and largest at the late layer. Fine-tuning the SAE on each generation largely restores reconstruction, suggesting that much of this degradation reflects a mismatch between the original SAE and the shifted activation distribution. However, recovering reconstruction does not necessarily recover feature-level usefulness. On \texttt{Pythia-160M}, sparse probes using only the top few latents remain degraded after fine-tuning even when probes over the full SAE representation recover, indicating that task-relevant information persists but becomes less concentrated in individual latents. On \texttt{Gemma-2-2B}, these feature-level effects vary across layers and metrics. Taken together, these results show that reconstruction quality and feature-level usefulness can diverge under self-consumption. Evaluating whether an adapted SAE remains a reliable interpretability tool therefore requires feature-level criteria in addition to reconstruction fidelity.

 The remainder of the paper is organized as follows: Sec.~\ref{sec:related} reviews related work, Sec.~\ref{sec:setup} introduces the matched-lineage setup and SAE evaluation regimes, and Sec.~\ref{sec:theory} develops a linear-autoencoder analysis that separates the reconstruction cost of a stale dictionary from the intrinsic reconstructability of the shifted representation, and shows why reconstruction recovery need not imply recovery of feature localization. Sections~\ref{sec:experiments}--\ref{sec:discussion} then present the experiments, results, and broader implications.

\section{Related Work}\label{sec:related}

\paragraph{Self-consuming generative models:} Recursive training on model-generated data can cause distributional degradation, or model collapse~\citep{shumailov2023curse,shumailov2024ai,alemohammad2024self}. This effect has been documented in image generators~\citep{hataya2023will,martinez2023combining} and language models, where recursive training can reduce linguistic diversity and downstream performance~\citep{briesch2023large,guo2024curious}. Theoretical work characterizes when iterative retraining remains stable~\citep{bertrand2024stability}, how collapse alters scaling laws~\citep{dohmatob2024tale}, and how errors accumulate in regression and simplified language-model settings~\citep{seddik2024bad}. Proposed mitigations include accumulating synthetic data alongside real data rather than replacing it~\citep{gerstgrasser2024model,kazdan2024collapse}, filtering or verifying synthetic samples~\citep{feng2024beyond}, and introducing self-correction into the generation loop~\citep{gillman2024self}. Existing work primarily characterizes changes in the generated distribution or downstream behavior; we instead examine how self-consumption changes internal representations and the features recovered from them.
% build on sparse coding~\citep{olshausen1997sparse} and
\paragraph{Sparse autoencoders:} SAEs are motivated by the superposition hypothesis, under which models may represent more features than they have available dimensions~\citep{elhage2022toy, olshausen1997sparse}. They decompose dense activations into an overcomplete sparse dictionary whose features are often individually interpretable~\citep{bricken2023monosemanticity,cunningham2024sparse}. SAEs have since been scaled to language models~\citep{templeton2024scaling,gao2025scaling} and released in open feature suites~\citep{lieberum2024gemma}. Several variants improve the sparsity--reconstruction trade-off, including gated~\citep{rajamanoharan2024improving}, TopK~\citep{gao2025scaling}, BatchTopK~\citep{bussmann2024batchtopk}, JumpReLU~\citep{rajamanoharan2024jumping}, and Matryoshka~\citep{bussmann2025matryoshka} SAEs, as well as end-to-end training objectives~\citep{braun2024identifying}. SAE features have also been used for circuit discovery~\citep{marks2025sparse} and automated interpretation~\citep{bills2023language,paulo2024automatically}. Most of this literature evaluates SAEs on a fixed underlying model; our setting instead asks what happens when the model itself evolves across generations.

\paragraph{Interpretability across changing models:} Recent work suggests that some interpretability structure can persist across model changes. SAEs trained on base models can transfer to chat-tuned variants~\citep{kissane2024saes, gupte2025transferability}, feature spaces can be partially shared across language models~\citep{lan2024sparse}, and SAE representations can remain stable under pruning~\citep{ gupte2026pruningmeetsinterpretabilitypreserving}. Crosscoders explicitly learn joint dictionaries for comparing related models~\citep{lindsey2024sparse, minder2025overcoming}, while mechanistic studies of fine-tuning find substantial reuse and refinement of existing mechanisms~\citep{jain2024mechanistically,prakash2024fine}. More general representation-similarity methods such as SVCCA and CKA compare activations across models or training runs~\citep{raghu2017svcca,kornblith2019similarity}. These settings typically examine a single transition between models. We instead study a sequence of recursive updates and ask whether SAE transfer degrades progressively as those changes accumulate.

\paragraph{Evaluating SAEs beyond reconstruction:} Reconstruction fidelity alone does not determine whether an SAE has learned useful or interpretable features. Individual features can split or absorb one another~\citep{chanin2024absorption}, reconstruction error can contain structured ``dark matter''~\citep{engels2024decomposing}, different training seeds can produce different feature dictionaries~\citep{paulo2025sparse}, and SAEs trained on randomly initialized transformers can perform surprisingly well on some standard metrics~\citep{heap2025sparse}. These limitations have motivated evaluations based on supervised dictionaries~\citep{makelov2024towards}, sparse probing~\citep{gurnee2023finding,kantamneni2025sparse}, steering benchmarks~\citep{wu2025axbench}, and broader evaluation suites such as SAEBench~\citep{karvonen2025saebench}. We adopt this broader perspective to distinguish recovery of reconstruction from recovery of feature-level usefulness.

\paragraph{Linear autoencoders and spectral perturbation.} Our analysis in Sec.~\ref{sec:theory} builds on the equivalence between optimal linear autoencoders and principal component analysis~\citep{bourlard1988auto,baldi1989neural,kunin2019loss}. We combine this connection with eigenspace perturbation bounds~\citep{davis1970rotation} and finite-sample covariance concentration~\citep{vershynin2018high} to analyze how repeated shifts in the activation covariance make a fixed dictionary increasingly mismatched to the evolving representation.

\section{Problem Setup}\label{sec:setup}

\subsection{Self-Consuming Training and a Matched Real-Data Control}
 
Let $p^{(0)}$ be a language model pretrained on real data, and let $\mathcal{A}(p;\mathcal{D})$ denote continued next-token-prediction training of model $p$ on dataset $\mathcal{D}$ with a fixed optimization procedure and token budget. A self-consuming loop runs for $T$ generations starting from $p_{\syn}^{(0)} = p^{(0)}$. At generation $t \in \{1,\dots,T\}$, the previous model generates a synthetic corpus on which it is then trained:
\begin{equation}\label{eq:syn}
    \tilde{\mathcal{D}}^{(t)} = \{x_j^{(t)}\}_{j=1}^{n}, \quad x_j^{(t)} \overset{\text{i.i.d.}}{\sim} p_{\syn}^{(t-1)},
    \qquad
    p_{\syn}^{(t)} = \mathcal{A}\big(p_{\syn}^{(t-1)};\, \tilde{\mathcal{D}}^{(t)}\big).
\end{equation}
Comparing generation $t$ with generation $0$ would conflate the effect of synthetic data with the change caused by any continued training. We therefore construct a control lineage that starts from the same model and uses the same token budget and optimization procedure, but is trained on a fresh slice of real data at every generation:
\begin{equation}\label{eq:real}
    p_{\real}^{(t)} = \mathcal{A}\big(p_{\real}^{(t-1)};\, \mathcal{D}_{\real}^{(t)}\big),
    \qquad
    p_{\real}^{(0)} = p_{\syn}^{(0)} = p^{(0)}.
\end{equation}
The two lineages thus differ only in whether the continued-training data are generated by the previous model or drawn from real text.
 
\subsection{Zero-Shot and Fine-tuned SAEs}\label{sec:regimes}
 
Even with this control, a synthetic--real gap can arise for two different reasons: self-consumption may shift the activation distribution away from the one the SAE was trained on, leaving a stale dictionary, or it may change the geometry of the representation the dictionary is trying to describe. We separate these with two evaluation regimes. Let $S^{(0)}$ be an SAE trained on activations of $p^{(0)}$, and let $h_\ell^{(t)}$ denote activations of $p_\ell^{(t)}$ for lineage $\ell \in \{\syn,\real\}$.
 
\paragraph{Zero-shot (frozen) SAE.} We apply $S^{(0)}$ unchanged to every generation, $S_\ell^{(t)} = S^{(0)}$. This measures zero-shot transfer of the original dictionary as the representation changes.
 
\paragraph{Fine-tuned SAE.} We initialize from $S^{(0)}$ and continue training on activations of the target model, $S_\ell^{(t)} = \mathrm{SAEFT}\big(S^{(0)};\, h_\ell^{(t)}\big)$, separately for each lineage and generation. The dictionary is adapted, not retrained from scratch.
 
The two regimes coincide at generation $0$. If fine-tuning closes the synthetic--real gap, the damage lies in a stale dictionary; if the gap persists, self-consumption has changed the representation in a way that adapting the dictionary cannot undo.
 
\subsection{Measuring the Effect of Self-Consumption}
 
Let $m(p, S)$ denote an SAE evaluation metric computed with SAE $S$ on activations of model $p$. At each generation $t$, we compare the synthetic lineage with the generation-matched control:
\begin{equation}\label{eq:damage}
    \mathrm{Damage}_m^{(t)} = s_m \Big[\, m\big(p_{\real}^{(t)}, S_{\real}^{(t)}\big) - m\big(p_{\syn}^{(t)}, S_{\syn}^{(t)}\big) \Big],
\end{equation}
where $s_m = +1$ if larger values of $m$ are better and $s_m = -1$ otherwise. Positive damage therefore always means that the synthetic lineage performs worse than matched continued training on real data, and zero means no degradation attributable to self-consumption. Since both lineages start from the same model and SAE, damage is zero at generation $0$ by construction.

\section{Analysis of SAE Staleness Under Representational Drift}
\label{sec:theory}

Our experiments distinguish two phenomena: a fixed SAE can become poorly matched to an evolving representation, while an SAE adapted to that representation can still reconstruct it well. We formalize this distinction using a linear autoencoder as an analytically tractable proxy for an SAE. The goal is to isolate when reconstruction degradation reflects a stale dictionary, and why recovering reconstruction need not recover feature interpretability. Full derivations are given in Appendix~\ref{app:theory}.

\paragraph{Linear autoencoders recover a principal subspace.}
Let $h_{\ell}^{(t)} \in \mathbb{R}^{d}$ denote centered activations at generation $t$ from lineage $\ell \in \{\mathrm{syn},\mathrm{real}\}$, with covariance $\Sigma_{\ell}^{(t)}$. Consider a rank-$k$ linear autoencoder with orthonormal decoder $W$ and projection $P=WW^\top$. Its reconstruction error is
\begin{equation}
    \mathcal{L}(P;\Sigma)
    = \mathbb{E}\|h-Ph\|_2^2
    = \operatorname{tr}\!\left[(I-P)\Sigma\right].
    \label{eq:linear_recon}
\end{equation}
The optimal rank-$k$ projection $P_{\ell}^{(t)}$ is the projection onto the top-$k$ eigenspace of $\Sigma_{\ell}^{(t)}$. Thus, changes in the activation covariance can change the subspace that is optimal for reconstruction.

\paragraph{Representational drift creates a staleness penalty.}
Let $P^{(0)}$ denote the projection learned at generation $0$. The zero-shot linear SAE uses this fixed projection at every later generation,
\begin{equation}
    \mathcal{L}^{\mathrm{zs}}_{\ell}(t)
    = \mathcal{L}\!\left(P^{(0)};\Sigma_{\ell}^{(t)}\right),
\end{equation}
whereas an ideally adapted SAE uses the current optimal projection,
\begin{equation}
    \mathcal{L}^{\mathrm{adapt}}_{\ell}(t)
    = \mathcal{L}\!\left(P_{\ell}^{(t)};\Sigma_{\ell}^{(t)}\right).
\end{equation}
Their difference,
\begin{equation}
    S_{\ell}(t)
    :=
    \mathcal{L}^{\mathrm{zs}}_{\ell}(t)
    -
    \mathcal{L}^{\mathrm{adapt}}_{\ell}(t)
    =
    \operatorname{tr}
    \left[
        \left(P_{\ell}^{(t)}-P^{(0)}\right)
        \Sigma_{\ell}^{(t)}
    \right]
    \geq 0,
    \label{eq:staleness}
\end{equation}
is the reconstruction cost attributable specifically to using a stale generation-$0$ dictionary.

This penalty is controlled by the misalignment between the original and current principal subspaces. Let $\Theta_{\ell}^{(t)}$ denote their principal angles and
let
\[
    \gamma_{\ell}^{(t)}
    =
    \lambda_k\!\left(\Sigma_{\ell}^{(t)}\right)
    -
    \lambda_{k+1}\!\left(\Sigma_{\ell}^{(t)}\right)
\]
be the relevant eigengap. Then
\begin{equation}
    \gamma_{\ell}^{(t)}
    \|\sin\Theta_{\ell}^{(t)}\|_F^2
    \leq
    S_{\ell}(t)
    \leq
    \left(
        \lambda_1\!\left(\Sigma_{\ell}^{(t)}\right)
        -
        \lambda_d\!\left(\Sigma_{\ell}^{(t)}\right)
    \right)
    \|\sin\Theta_{\ell}^{(t)}\|_F^2.
    \label{eq:staleness_angles}
\end{equation}
Hence a frozen dictionary incurs increasing reconstruction cost when the reconstruction-relevant subspace moves away from the one on which it was trained. Standard eigenspace perturbation results further relate this misalignment to cumulative covariance shift; we give the corresponding finite-sample bounds in Appendix~\ref{app:theory}.

\paragraph{Why fine-tuning closes the synthetic--real gap.}
The matched control makes this decomposition particularly useful. Define reconstruction damage under zero-shot and adapted evaluation as
\begin{align}
    D_{\mathrm{zs}}(t)
    &=
    \mathcal{L}^{\mathrm{zs}}_{\mathrm{syn}}(t)
    -
    \mathcal{L}^{\mathrm{zs}}_{\mathrm{real}}(t),\\
    D_{\mathrm{adapt}}(t)
    &=
    \mathcal{L}^{\mathrm{adapt}}_{\mathrm{syn}}(t)
    -
    \mathcal{L}^{\mathrm{adapt}}_{\mathrm{real}}(t).
\end{align}
Then
\begin{equation}
    D_{\mathrm{zs}}(t)-D_{\mathrm{adapt}}(t)
    =
    S_{\mathrm{syn}}(t)-S_{\mathrm{real}}(t).
    \label{eq:damage_decomposition}
\end{equation}
Thus, if self-consumption induces greater subspace drift than fine-tuning on real data, the synthetic lineage acquires a larger staleness penalty, and adapting the SAE removes this component of the synthetic--real gap precisely. This is the pattern observed in our reconstruction experiments.

\paragraph{Reconstruction recovery need not imply feature recovery.}
Crucially, reconstruction constrains a subspace more strongly than it constrains the individual coordinates within that subspace. Suppose two representations of the same $k$-dimensional subspace differ by an orthogonal change of basis,
$W' = WQ$. Then
\[
    W'W'^\top = WQQ^\top W^\top = WW^\top,
\]
so their reconstruction is identical. Their latent coordinates, however, satisfy $z'=Q^\top z$ and can distribute information very differently across individual dimensions. A linear probe using the full latent vector is invariant to this transformation: for every probe $a^\top z$ there is a probe $(Q^\top a)^\top z'$ with identical predictions. In contrast, information that was concentrated in one or a few coordinates can be spread across many coordinates after the change of basis.

This gives a simple explanation for why reconstruction and sparse localization can diverge: adaptation can recover a good representation of the relevant subspace without requiring task-relevant information to remain concentrated in the same small number of SAE latents. The argument predicts the possibility, rather than the necessity, of feature-level degradation, consistent with the strong effect in Pythia and the more heterogeneous behavior in Gemma.

\section{Experimental Setup}\label{sec:experiments}

\paragraph{Self-consuming lineages.}
We study five generations of synthetic and matched real-data training using \texttt{Pythia-160M}~\citep{biderman2023pythia} and \texttt{Gemma-2-2B}~\citep{gemmateam2024gemma}. Starting from the same pretrained checkpoint, each synthetic generation is trained exclusively on text generated by the previous synthetic model, while the real-data lineage is trained on a fresh slice of \nolinkurl{monology/pile-uncopyrighted}. Both lineages use the same training budget and optimization setup, isolating the effect of synthetic rather than continued training. We analyze Pythia layer 8 of 12, matching SAEBench, and Gemma layers 5, 12, and 19 to test how representational change varies with depth. Full training and sampling details are provided in Appendix~\ref{app:training}.

\paragraph{SAE configurations.}
For \texttt{Pythia-160M}, we vary both SAE width and architecture. Our width sweep uses BatchTopK~\citep{bussmann2024batchtopk} SAEs with widths $\{2^{12},2^{14},2^{16}\}$, while our architecture sweep fixes the width at $2^{14}$ and compares BatchTopK, JumpReLU~\citep{rajamanoharan2024jumping}, and Matryoshka-BatchTopK~\citep{bussmann2025matryoshka}. For \texttt{Gemma-2-2B}, we use width-$2^{14}$ JumpReLU SAEs at each evaluated layer. Every SAE is evaluated under both regimes from Sec.~\ref{sec:regimes}. For fine-tuning, we use three seeds and report their mean and spread.

\paragraph{Evaluation metrics.}
We use four SAEBench~\citep{karvonen2025saebench} evaluation families to separate reconstruction quality from feature-level usefulness (Appendix~\ref{app:metrics}). \textit{Core metrics} measure sparsity ($L_0$), reconstruction MSE, and explained variance. We complement these with \textit{sparse probing}, which measures how well task-relevant concepts can be recovered from a small number of SAE latents~\citep{gurnee2023finding,kantamneni2025sparse}. We train probes using the top $k\in\{1,2,5\}$ latents and the full SAE representation. Comparing top-$k$ and full-SAE probing distinguishes loss of information from loss of sparse localization: if top-$k$ performance degrades while full-SAE performance remains stable, the information remains present but is distributed across more latents.

Finally, we use two \textit{causal ablation} benchmarks: spurious correlation removal (SCR) and targeted probe perturbation (TPP). We treat sparse probing as our primary measure of feature localization, with SCR and TPP providing complementary evidence about causal feature usefulness.

\section{Results}\label{sec:results}
% \TODO{Review starting here}

Unless otherwise stated, we report results as damage relative to the generation-matched real-data control (Eq.~\ref{eq:damage}), so that positive values indicate worse performance under self-consumption.

% OPTIONAL: main-text perplexity figure. It costs ~0.25 page. Enable it only if
% Section 4 is <= ~1.2 pages; also change the reference in Sec. 6.1 below to
% Fig.~\ref{fig:perplexity}.
% \begin{figure}[t]
%     \centering
%     \includegraphics[width=0.75\linewidth]{figures/perplexity_main.pdf}
%     \caption{\textbf{Self-consumption degrades language modeling.} Perplexity on held-out real data. The synthetic lineage worsens every generation; the real-data control plateaus (Pythia) or improves (Gemma).}
%     \label{fig:perplexity}
% \end{figure}
 
\begin{figure}[t]
    \centering
    \includegraphics[width=\linewidth]{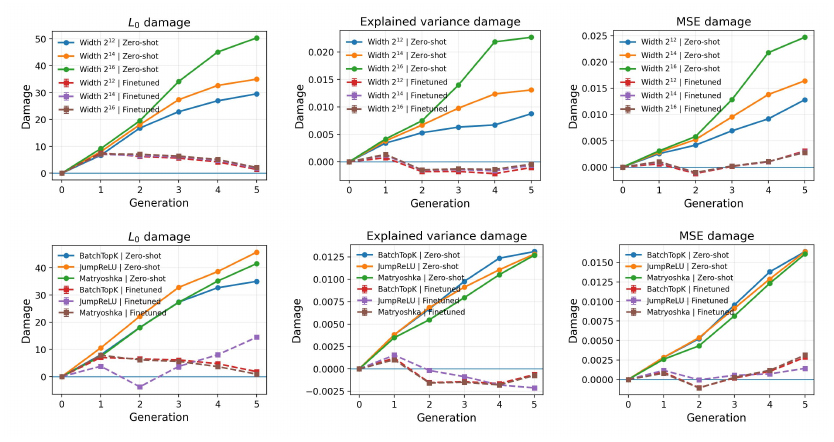}
    \caption{\textbf{Zero-shot damage to \texttt{Pythia-160M} SAEs grows with self-consumption; fine-tuning removes most of it.} Top: BatchTopK SAEs with widths $2^{12}$, $2^{14}$, and $2^{16}$. Bottom: BatchTopK, JumpReLU, and Matryoshka-BatchTopK SAEs at width $2^{14}$. Solid lines denote zero-shot evaluation with the original SAE held fixed; dashed lines denote SAEs fine-tuned separately on each generation (error bars show variation across fine-tuning seeds). All quantities are damage relative to the matched real-data lineage; higher values indicate greater degradation under self-consumption.}
    % \vspace{-10px}
    \label{fig:pythia_core}
\end{figure}

\paragraph{Self-Consumption Progressively Degrades the Underlying Models.} We first establish that the self-consuming lineages exhibit cumulative behavioral degradation. As shown in Appendix \ref{app:perplexity}, Fig.~\ref {fig:perplexity_gap}, perplexity on held-out real text increases across generations for both synthetic lineages. By generation 5, perplexity is $2.13\times$ that of the matched control for \texttt{Pythia-160M} and $1.46\times$ for \texttt{Gemma-2-2B}, corresponding to log-perplexity gaps of $0.75$ and $0.38$ nats, respectively. By comparison, continued training on real data produces much smaller changes in perplexity. Pythia's real-data lineage rises from a perplexity of $23.2$ to roughly $30$ after the first few generations and then plateaus, while Gemma's control improves with continued real-data training. Thus, the widening synthetic--real gap cannot be explained by continued training alone: recursive training on model-generated data introduces additional degradation that accumulates across generations. Having established this behavioral divergence, we next ask whether self-consumption produces a corresponding shift in the internal representations captured by SAEs.

\paragraph{Pythia: Zero-Shot SAEs Become Progressively Stale.} The solid lines in Fig.~\ref{fig:pythia_core} show core-metric damage when the generation-0 Pythia SAE is applied zero-shot to later checkpoints. Damage increases steadily across generations in $L_0$, explained variance, and MSE for every configuration. Across widths, the widest dictionary exhibits the largest degradation; across architectures, BatchTopK, JumpReLU, and Matryoshka-BatchTopK follow qualitatively similar trends, indicating that the effect is not specific to a particular SAE design. Because each synthetic checkpoint is compared against the real-data control at the same generation, these trends isolate the additional representational shift induced by self-consumption. This progressive degradation indicates that the original SAE becomes increasingly misaligned to the activation distribution induced by self-consumption.
 
 \begin{figure}[t]
    \centering
    \includegraphics[width=\linewidth]{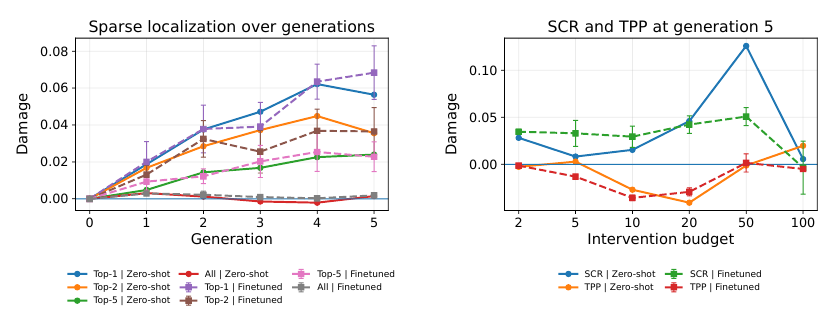}
    \caption{\textbf{Fine-tuning does not restore sparse localization in \texttt{Pythia-160M}.} Left: sparse-probing damage using the top-$k$ latents or the full SAE representation. Damage remains high for small $k$ after fine-tuning, while full-SAE probing stays near the real-data control, indicating that task-relevant information persists but is less concentrated in individual latents. Right: SCR/TPP damage at gen. 5 across intervention budgets, providing complementary feature-level evaluations.}
    % \vspace{-10px}
    \label{fig:pythia_feature}
\end{figure}

\paragraph{Pythia: Fine-tuning Restores Reconstruction but Not Sparse Localization.}\label{sec:pythia_ft}
The dashed lines in Fig.~\ref{fig:pythia_core} show what happens when the original SAE is adapted to each generation. Fine-tuning removes most of the synthetic--real gap in explained variance and MSE and substantially reduces the $L_0$ gap. Residual sparsity damage remains, most notably for JumpReLU, but is much smaller than under zero-shot transfer. This recovery indicates that the shifted activations remain well captured by a sparse dictionary once the SAE is allowed to adapt, and that much of the zero-shot degradation arises from mismatch between the original dictionary and the evolving representation. Reconstruction recovery, however, does not establish that task-relevant information remains localized in the same way across individual SAE latents.

 Feature-level behavior follows a different pattern (Fig.~\ref{fig:pythia_feature}, left). Sparse-probing damage increases across generations when probes are restricted to the top few latents, with the largest effect for top-$1$ probing, and fine-tuning does not eliminate this degradation. In contrast, probes over the full SAE representation show little to no damage in either regime. This divergence suggests that task-relevant information is not lost from the SAE representation, but becomes less concentrated in a small number of individual latents. Self-consumption therefore reduces the sparse localization of task-relevant information even after reconstruction quality has largely recovered. The intervention-based metrics show a less uniform pattern (Fig.~\ref{fig:pythia_feature}, right). At generation 5, SCR damage is positive across most intervention budgets and is not consistently removed by fine-tuning, whereas TPP damage remains near zero or changes sign. Thus, feature-level degradation is evident in sparse localization and some causal metrics, but does not appear uniformly across all measures of SAE feature quality.

\begin{figure}[t]
    \centering
    \includegraphics[width=\linewidth]{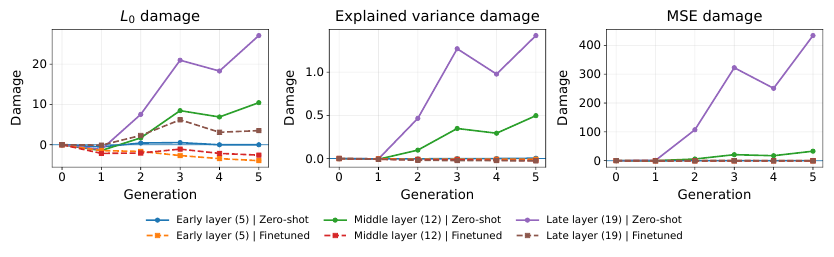}\\[2pt]
    \includegraphics[width=\linewidth]{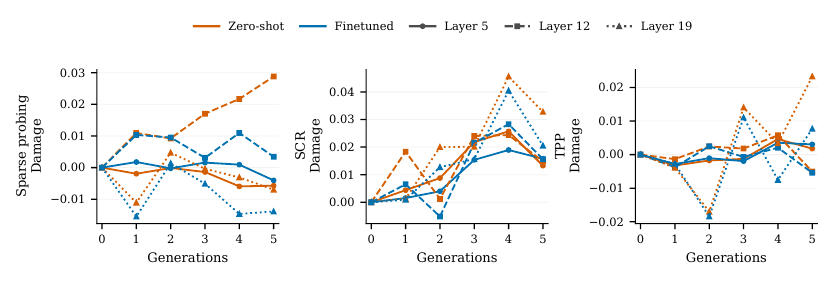}
    \caption{\textbf{Self-consumption damage in \texttt{Gemma-2-2B} SAEs at an early (5), middle (12), and late (19) layer.} Top: core metrics (solid: zero-shot; dashed: fine-tuned). Zero-shot damage grows with depth and generation; fine-tuning removes nearly all of it, leaving a small residual $L_0$ gap at layer 19. Bottom: sparse-probing, SCR, and TPP damage; color denotes the regime (orange: zero-shot; blue: fine-tuned) and line style the layer. Higher is worse.}
    % \vspace{-10px}
    \label{fig:gemma}
\end{figure}
 
\paragraph{Gemma: Representational Drift Is Strongly Depth-Dependent.} Fig.~\ref{fig:gemma} (top) shows that zero-shot SAE degradation in \texttt{Gemma-2-2B} depends strongly on depth. The early layer remains largely stable across generations, the middle layer develops a moderate synthetic--real gap, and the late layer shows large, progressively growing damage in $L_0$, explained variance, and MSE. Fine-tuning, however, nearly eliminates the explained-variance and MSE gaps at all three layers, including the late layer, while leaving only a small residual sparsity gap. Thus, self-consumption produces substantially greater representational drift at later layers, but these shifted representations remain aligned to sparse reconstruction once the SAE is adapted.
 
\paragraph{Gemma: Feature-Level Degradation Is Not Consistent Across Metrics.} Gemma's feature-level behavior is less uniform than its reconstruction-level drift (Fig.~\ref{fig:gemma}, bottom). For sparse probing, the clearest sustained zero-shot degradation appears at the middle layer, where damage grows over later generations but is substantially reduced by fine-tuning. The early layer remains largely stable, while the late layer fluctuates without a consistent trend. Unlike Pythia, Gemma therefore shows no robust loss of sparse localization after SAE adaptation. The intervention-based metrics are similarly mixed. SCR damage emerges in later generations and is generally larger at deeper layers; fine-tuning reduces this damage but does not consistently eliminate it. TPP shows no clear pattern, with damage changing sign across generations and layers and adaptation sometimes helping and sometimes hurting. Overall, the strong depth-dependent drift seen in reconstruction does not translate into equally strong or persistent degradation across feature-level interpretability metrics.
\section{Discussion and Limitations}\label{sec:discussion}

\paragraph{Self-consumption makes fixed SAE dictionaries stale.}
Across both model families, SAEs trained before self-consumption transfer progressively worse to later generations. Because each synthetic checkpoint is compared with a generation-matched real-data control, this degradation isolates the effect of synthetic rather than continued training itself. Yet fine-tuning the original SAE removes most of the reconstruction gap, even where zero-shot failure is largest. The representations therefore remain amenable to sparse reconstruction; the main failure is that the original dictionary becomes misaligned with the shifted activation distribution. In practice, an SAE that degrades after model updates may need adaptation rather than replacement.

\paragraph{Representational drift grows with depth.}
In Gemma, zero-shot degradation is small at an early layer and much larger deeper in the network. This is consistent with prior analyses of Gemma Scope SAEs: early-layer features often track token-level structure, while deeper features capture more conceptual and context-dependent information~\citep{sun2025dense}. Structural properties may be relatively stable across human- and model-generated text, whereas deeper semantic representations may be more sensitive to the distributional narrowing induced by self-consumption. Our experiments establish the depth dependence but do not identify its mechanism. Importantly, the largest drift appears in the deeper layers that are often most interesting for interpretability.

\paragraph{Reconstruction recovery does not guarantee feature recovery.} Pythia provides the clearest example. After fine-tuning, reconstruction largely recovers, but sparse probing with the top few latents remains degraded while probing the full SAE representation does not. Task-relevant information therefore remains accessible but becomes less concentrated in individual features. Gemma shows a more heterogeneous pattern: sparse probing is largely repaired, SCR retains residual degradation, and TPP shows no consistent direction. These metrics capture different properties: localization versus causal usefulness, and need not move together. More broadly, reconstruction metrics can reveal when an SAE has become stale, but successful reconstruction after adaptation is not sufficient to establish that its features remain equally useful for interpretation.

\paragraph{Limitations.} Our experiments are intentionally controlled rather than exhaustive. We study two model families over five generations to test whether the same pattern appears across different scales and architectures while keeping the recursive training process tractable. We use pure synthetic training, without mixing in real data, because this isolates the effect of self-consumption; real--synthetic mixtures are more realistic, but would make it harder to attribute changes specifically to synthetic data. We analyze one Pythia layer and early, middle, and late Gemma layers rather than every layer, which lets us test both robustness across SAE designs and variation with depth without greatly expanding the experiment. We also fine-tune the original SAE rather than train a new SAE from scratch because our main question is whether an existing interpretability tool remains reliable as the model evolves. Likewise, we use standardized SAEBench metrics so that reconstruction, localization, and causal behavior can be compared consistently across generations, although these metrics do not capture every aspect of semantic interpretability. Finally, each model lineage uses one training seed because independently training multiple five-generation lineages is computationally expensive; our reported error bars therefore capture variation in SAE fine-tuning rather than variation across model-training runs. Larger models, additional lineage seeds, broader layer coverage, and more realistic mixtures of real and synthetic data are natural next steps.

\section{Conclusion}\label{sec:conclusion}

Through this work, we studied how self-consumption changes the internal representations exposed by sparse autoencoders. Across both model families, fixed SAEs become progressively less reliable under self-consumption, with the strongest degradation appearing in later layers. Yet fine-tuning largely restores reconstruction, showing that the representations remain compressible even as the original SAE dictionary becomes stale. The key finding is that reconstruction recovery does not imply feature recovery. In Pythia, task-relevant information remains accessible in the full SAE representation while becoming less concentrated in individual latents; in Gemma, feature-level effects depend on layer and metric. These results highlight an important distinction for interpretability on evolving models: representation shift, SAE adaptation, and feature-level usefulness are separate questions. Reconstruction can tell us whether an SAE still fits the activations, but not whether its features remain equally useful for interpretation.

\

\subsection*{AI use statement}
In this work, we used generative AI tools to assist with writing, code generation, and verification. For writing, we provided the models with raw notes, arguments, and unfiltered draft content and used them to improve clarity, organization, and presentation, including refining portions of the introduction and polishing the manuscript. For visualization, we provided experimental results in CSV format together with the intended comparisons and plot designs, and used generative AI to produce and iteratively refine plotting code and figure formatting for the ICLR style. We also used generative AI as an additional check of the theoretical analysis, including reviewing derivations for mathematical consistency and identifying potential issues.

We did not use generative AI to generate experimental results or fabricate data. All AI-assisted text was reviewed and edited by the authors; plots were checked against the underlying experimental values; and the theoretical arguments and derivations were independently reviewed for correctness by the authors. We take responsibility for the final content of this work, including text, claims, code, figures, and other artifacts produced with the aid of generative AI.

\bibliography{iclr2027_conference}
\bibliographystyle{iclr2027_conference}

\appendix
\section*{Appendix} \label{app}
\section{Experimental Details}
\subsection{Training and Sampling Details}\label{app:training}
 
\paragraph{Real data.} Real training data are streamed from \texttt{monology/pile-uncopyrighted} and shuffled with a buffer of 10{,}000 documents. Documents are tokenized, an end-of-sequence token is appended to each, and the resulting token stream is split into 512-token blocks. The first 500 blocks form a fixed held-out evaluation set shared by both lineages. The real-data lineage then consumes consecutive, non-overlapping slices of 10{,}000 blocks, so no real block is reused across generations or overlaps with the evaluation set. Because Pythia was pretrained on the Pile, these blocks are new to the fine-tuning loop but not necessarily new to the model.
 
\paragraph{Synthetic data.} At each generation of the synthetic lineage, all training data are sampled from the previous-generation checkpoint, with no real data mixed in. Each sample starts from a single beginning-of-sequence token and is generated for up to 512 new tokens with temperature $1.0$ and top-$p = 1.0$. We explicitly disable top-$k$ sampling so that default generation configurations cannot truncate the tail of the distribution. The prompt token is discarded, each sample is truncated after its first end-of-sequence token, and the samples are concatenated and split into 512-token blocks, mirroring the real-data packing.
 
\paragraph{Optimization.} Each generation is fine-tuned for one epoch on its 10{,}000 blocks using the settings in Table~\ref{tab:hparams}. Settings are identical across the two lineages of each model and differ between models only where noted. Pythia is trained in mixed precision with fp32 master weights. Gemma is trained entirely in bfloat16 with gradient checkpointing. Because bfloat16 master weights can round very small updates to zero, Gemma's effective step size may be smaller than its nominal learning rate, so cross-model comparisons of effect magnitude should be interpreted with caution.
 
\begin{table}[t]
    \centering
    \caption{Fine-tuning and sampling settings. Each setting is shared by the real and synthetic lineages of a given model.}
    \label{tab:hparams}
    \small
    \begin{tabular}{lll}
        \toprule
        & \textbf{\texttt{Pythia-160M}} & \textbf{Gemma-2-2B} \\
        \midrule
        Checkpoint & \texttt{pythia-160m-deduped} & \texttt{gemma-2-2b} \\
        Blocks per generation & 10{,}000 & 10{,}000 \\
        Block length (tokens) & 512 & 512 \\
        Tokens per generation & $\approx$5.1M & $\approx$5.1M \\
        Epochs per generation & 1 & 1 \\
        Optimizer & AdamW & AdamW \\
        Learning rate & $5 \times 10^{-5}$ & $2 \times 10^{-5}$ \\
        LR schedule & linear, 3\% warmup & linear, 3\% warmup \\
        Weight decay & 0 & 0 \\
        Gradient clipping (max norm) & 1.0 & 1.0 \\
        Per-device batch size & 16 & 4 \\
        Gradient accumulation steps & 2 & 8 \\
        Effective batch size & 32 & 32 \\
        Precision & fp32 weights, bf16 autocast & bf16 throughout \\
        Gradient checkpointing & No & Yes \\
        Sampling & $T = 1.0$, top-$p = 1.0$, no top-$k$ & $T = 1.0$, top-$p = 1.0$, no top-$k$ \\
        Held-out evaluation blocks & 500 & 500 \\
        Random seed & 0 & 0 \\
        SAE layers & 8 & 5, 12, 19 \\
        Generations analyzed & 0--5 & 0--5 \\
        \bottomrule
    \end{tabular}
\end{table}
\subsection{Evaluation Metric Details}\label{app:metrics}
 
All metrics are computed with SAEBench~\citep{karvonen2025saebench} at the SAE layers listed in Table~\ref{tab:hparams}. Following SAEBench, BOS, EOS, and padding tokens are excluded from all evaluations.
 
\paragraph{Core metrics.} Core metrics are computed on 128-token OpenWebText sequences, using 3{,}200 sequences for the loss-based scores and 32{,}000 for sparsity statistics. $L_0$ is the mean number of active latents per token. Reconstruction MSE is the mean squared error between an activation $\vx$ and its SAE reconstruction $\hat{\vx}$. Explained variance is the fraction of the variance of $\vx$ captured by $\hat{\vx}$.
 
\paragraph{Sparse probing.} Sparse probing uses 35 one-versus-all binary classification tasks drawn from five datasets: profession classification (Bias in Bios), product category and sentiment classification (Amazon Reviews), language identification (Europarl), programming-language identification (GitHub code), and news topic classification (AG News). Each task uses 4{,}000 training and 1{,}000 test examples, truncated to 128 tokens. Inputs are passed through the model and encoded by the SAE, and latent activations are mean-pooled over non-padding tokens. The top $k$ latents are selected by the difference in their mean activation between the positive and negative classes, and a logistic-regression probe is trained on them and evaluated on the held-out test examples. We use $k \in \{1, 2, 5\}$ and also train a probe on all SAE latents. The residual-stream baseline $\mathrm{Acc}_{\mathrm{LLM},k}$ applies the same pooling, selection, and probing procedure to the residual-stream dimensions at the same layer, so $\mathrm{SPGap}_1$ compares probes of equal dimension. Reported values are test accuracies averaged over tasks.
 
\paragraph{Latent selection and ablation for SCR and TPP.} Both evaluations select latents by their attribution to a linear probe. The attribution score combines the projection of a latent's decoder direction onto the probe with the difference in the latent's mean activation between inputs with and without the probed concept. Selected latents are zero-ablated, and the original SAE reconstruction error is added back to the modified reconstruction so that the intervention changes only the targeted latents. For both SCR and TPP, we evaluate intervention budgets of $k \in \{2, 5, 10, 20, 50, 100\}$ ablated latents.
 
\paragraph{Spurious correlation removal (SCR).} SCR adapts the SHIFT procedure of \citet{marks2025sparse}. Each task is built from a dataset with an intended label and a spurious label: profession and gender in Bias in Bios, and product category and sentiment in Amazon Reviews. The data are split into a biased subset, in which the two labels are perfectly correlated (for example, only male professors and female nurses), and a balanced subset containing all four combinations. A linear probe trained on the biased subset relies on both signals. We zero-ablate the $k$ latents with the highest attribution to a probe trained on the spurious label and measure the biased probe's accuracy on the intended label over the balanced subset. The score is
\begin{equation}
    S_{\mathrm{SCR}} = \frac{A_{\mathrm{abl}} - A_{\mathrm{base}}}{A_{\mathrm{oracle}} - A_{\mathrm{base}}},
\end{equation}
where $A_{\mathrm{base}}$ and $A_{\mathrm{abl}}$ are the biased probe's accuracies before and after ablation, and $A_{\mathrm{oracle}}$ is the accuracy of a probe trained directly on the intended label. The biased probe is not retrained after ablation. A score of 1 means the ablation recovers all of the accuracy lost to the spurious correlation, and 0 means it recovers none.
 
\paragraph{Targeted probe perturbation (TPP).} TPP extends this idea to multiclass datasets without requiring correlated labels. For each class $c_j$, we train a one-versus-rest linear probe with accuracy $A_j$. For each class $c_i$, we select the $k$ latents with the highest signed attribution to that class's probe, and let $A_{i,j}$ denote the accuracy of the probe for $c_j$ after the latents selected for $c_i$ are zero-ablated. The score is
\begin{equation}
    S_{\mathrm{TPP}} = \operatorname*{mean}_{i=j}\big(A_j - A_{i,j}\big) - \operatorname*{mean}_{i \neq j}\big(A_j - A_{i,j}\big),
\end{equation}
that is, the mean accuracy drop of each targeted probe minus the mean drop of the non-targeted probes. A high score means that the latents selected for a class damage that class's probe while leaving the other probes largely intact.
\subsection{Additional Language-Modeling Results}\label{app:perplexity}
 
\begin{figure}[!ht]
    \centering
    \includegraphics[width=\linewidth]{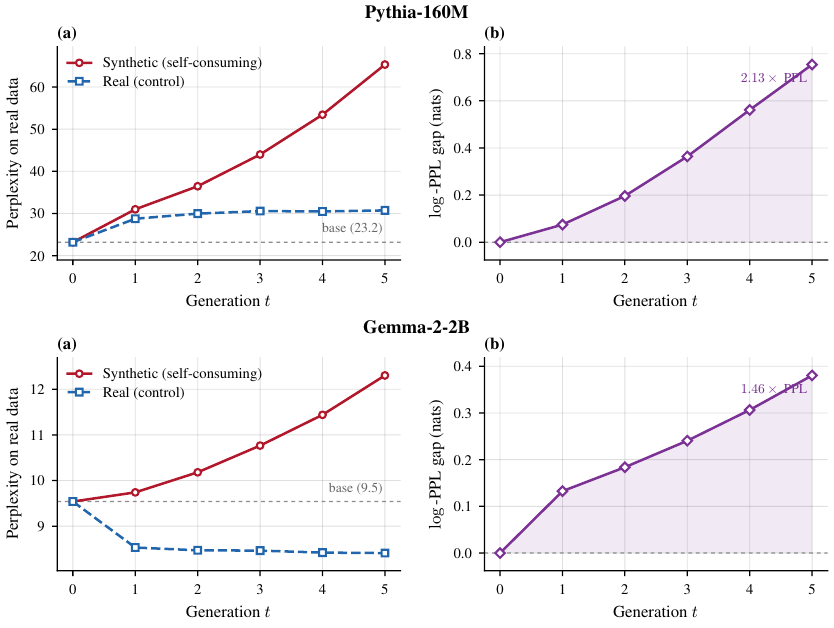}
    \caption{\textbf{Self-consumption degrades language modeling.} (a) Perplexity on held-out real data for the synthetic lineage and the matched real-data control. (b) Log-perplexity gap between the two lineages. By generation 5, the gap reaches $0.75$ nats ($2.13\times$ perplexity) for \texttt{Pythia-160M} and $0.38$ nats ($1.46\times$) for \texttt{Gemma-2-2B}. The control plateaus after the first generation for \texttt{Pythia-160M} and improves for \texttt{Gemma-2-2B}.}
    \label{fig:perplexity_gap}
\end{figure}

As a sanity check that the recursive training procedure produces the expected behavioral degradation, we evaluate each generation on held-out real text. Figure~\ref{fig:perplexity_gap} shows that perplexity increases progressively along the synthetic lineage relative to the generation-matched real-data control for both model families. By generation 5, the synthetic--real log-perplexity gap reaches $0.75$ nats for \texttt{Pythia-160M} and $0.38$ nats for \texttt{Gemma-2-2B}, corresponding to $2.13\times$ and $1.46\times$ higher perplexity, respectively. In contrast, continued training on real data produces substantially smaller changes. These results confirm that the self-consuming lineages accumulate degradation beyond that induced by continued training alone, providing the behavioral context for the representational shifts analyzed in the main text.

\section{Theoretical Details} \label{app:theory}

This appendix provides the derivations underlying the linear-autoencoder analysis in Sec.~\ref{sec:theory}. The analysis is intended as an analytically tractable model of SAE staleness rather than an exact model of nonlinear SAE training. Its main purpose is to separate three quantities: (i) how well the current representation can be reconstructed by an adapted dictionary, (ii) the additional reconstruction cost incurred by keeping an old dictionary fixed, and (iii) the localization of information within individual latent coordinates.

\subsection{Linear Autoencoders Recover a Principal Subspace}
\label{app:linear_pca}

Let $h\in\mathbb{R}^d$ denote a zero-mean activation vector with covariance
\[
    \Sigma = \mathbb{E}[hh^\top],
\]
and eigendecomposition
\[
    \Sigma = V\Lambda V^\top,
    \qquad
    \lambda_1\geq\lambda_2\geq\cdots\geq\lambda_d\geq 0.
\]

Consider a rank-$k$ linear autoencoder with decoder $W\in\mathbb{R}^{d\times k}$ satisfying $W^\top W=I_k$ and tied encoder $W^\top$. Its reconstruction is
\[
    \hat h = WW^\top h.
\]
Writing $P=WW^\top$, the expected reconstruction loss is
\begin{align}
    \mathcal{L}(P;\Sigma)
    &=\mathbb{E}\|h-Ph\|_2^2 \\
    &=\operatorname{tr}\!\left[(I-P)\Sigma\right] \\
    &=\operatorname{tr}(\Sigma)-\operatorname{tr}(P\Sigma).
    \label{eq:app_linear_loss}
\end{align}

\begin{proposition}[Linear autoencoder optimum]
\label{prop:linear_pca}
Among all rank-$k$ orthogonal projections, the reconstruction loss in Eq.~\ref{eq:app_linear_loss} is minimized by projection onto the top-$k$
eigenspace of $\Sigma$:
\[
    P^\star = V_kV_k^\top,
\]
where $V_k=[v_1,\ldots,v_k]$. The corresponding minimum reconstruction error is
\[
    \mathcal{L}^\star(\Sigma)
    =
    \sum_{i=k+1}^{d}\lambda_i.
\]
\end{proposition}

\begin{proof}
Since $\operatorname{tr}(\Sigma)$ is independent of $P$, minimizing Eq.~\ref{eq:app_linear_loss} is equivalent to maximizing $\operatorname{tr}(P\Sigma)$. By the Ky Fan maximum principle, the maximizing rank-$k$ projection is the projection onto the eigenvectors corresponding to the $k$ largest eigenvalues. Hence
\[
    \operatorname{tr}(P^\star\Sigma)
    =
    \sum_{i=1}^{k}\lambda_i,
\]
and therefore
\[
    \mathcal{L}^\star
    =
    \sum_{i=1}^{d}\lambda_i
    -
    \sum_{i=1}^{k}\lambda_i
    =
    \sum_{i=k+1}^{d}\lambda_i.
\]
\end{proof}

Thus, in the linear setting, adapting an autoencoder to a new activation distribution corresponds to adapting the subspace used for reconstruction.

\subsection{Dictionary Staleness Under Subspace Drift}
\label{app:staleness}

Let $\Sigma_{\ell}^{(t)}$ denote the activation covariance at generation $t$ for lineage $\ell\in\{\mathrm{syn},\mathrm{real}\}$, and let
\[
    P_{\ell}^{(t)}
\]
denote its optimal rank-$k$ projection. Both lineages share the same generation-$0$ covariance and therefore the same initial projection
\[
    P^{(0)}
    =
    P_{\mathrm{syn}}^{(0)}
    =
    P_{\mathrm{real}}^{(0)}.
\]

A frozen generation-$0$ autoencoder incurs reconstruction error
\[
    \mathcal{L}^{\mathrm{zs}}_{\ell}(t)
    =
    \mathcal{L}\!\left(P^{(0)};\Sigma_{\ell}^{(t)}\right),
\]
whereas an ideally adapted autoencoder obtains
\[
    \mathcal{L}^{\mathrm{adapt}}_{\ell}(t)
    =
    \mathcal{L}\!\left(P_{\ell}^{(t)};
                       \Sigma_{\ell}^{(t)}\right).
\]

We define the \emph{staleness penalty}
\begin{align}
    S_{\ell}(t)
    &:=
    \mathcal{L}^{\mathrm{zs}}_{\ell}(t)
    -
    \mathcal{L}^{\mathrm{adapt}}_{\ell}(t)\\
    &=
    \operatorname{tr}
    \left[
        \left(
            P_{\ell}^{(t)}-P^{(0)}
        \right)
        \Sigma_{\ell}^{(t)}
    \right].
    \label{eq:app_staleness}
\end{align}
Because $P_{\ell}^{(t)}$ minimizes reconstruction error for
$\Sigma_{\ell}^{(t)}$,
\[
    S_{\ell}(t)\geq 0.
\]

The quantity therefore measures exactly how much reconstruction is lost
because the dictionary is held fixed rather than adapted to the current
representation.

\subsubsection{Staleness is controlled by subspace misalignment}

Let the eigenvalues of $\Sigma_{\ell}^{(t)}$ be
\[
    \lambda_1^{(t)}
    \geq\cdots\geq
    \lambda_d^{(t)},
\]
and define the eigengap at the reconstruction boundary as
\[
    \gamma_{\ell}^{(t)}
    =
    \lambda_k^{(t)}
    -
    \lambda_{k+1}^{(t)}.
\]

Let
\[
    \Theta_{\ell}^{(t)}
\]
denote the principal angles between the range of $P^{(0)}$ and the
current optimal subspace $P_{\ell}^{(t)}$.

\begin{proposition}[Staleness and principal angles]
\label{prop:staleness_angles}
If $\gamma_{\ell}^{(t)}>0$, then
\begin{equation}
    \gamma_{\ell}^{(t)}
    \left\|
        \sin\Theta_{\ell}^{(t)}
    \right\|_F^2
    \leq
    S_{\ell}(t)
    \leq
    \left(
        \lambda_1^{(t)}-\lambda_d^{(t)}
    \right)
    \left\|
        \sin\Theta_{\ell}^{(t)}
    \right\|_F^2.
    \label{eq:app_staleness_angle_bound}
\end{equation}
\end{proposition}

\begin{proof}
Let $u_1,\ldots,u_d$ denote the eigenvectors of
$\Sigma_{\ell}^{(t)}$ and define
\[
    a_i=u_i^\top P^{(0)}u_i.
\]
Since $P^{(0)}$ is a rank-$k$ projection,
$0\leq a_i\leq1$ and
\[
    \sum_{i=1}^{d}a_i=k.
\]
The staleness penalty can be written as
\begin{align}
    S_{\ell}(t)
    &=
    \sum_{i=1}^{k}\lambda_i^{(t)}
    -
    \sum_{i=1}^{d}\lambda_i^{(t)}a_i\\
    &=
    \sum_{i=1}^{k}
        \lambda_i^{(t)}(1-a_i)
    -
    \sum_{i=k+1}^{d}
        \lambda_i^{(t)}a_i.
\end{align}

Moreover,
\[
    s:=
    \sum_{i=1}^{k}(1-a_i)
    =
    \sum_{i=k+1}^{d}a_i
    =
    \|\sin\Theta_{\ell}^{(t)}\|_F^2.
\]

Using
$\lambda_i^{(t)}\geq\lambda_k^{(t)}$ for $i\leq k$
and
$\lambda_i^{(t)}\leq\lambda_{k+1}^{(t)}$ for $i>k$,
\[
    S_{\ell}(t)
    \geq
    \left(
        \lambda_k^{(t)}
        -
        \lambda_{k+1}^{(t)}
    \right)s.
\]
Similarly,
\[
    S_{\ell}(t)
    \leq
    \left(
        \lambda_1^{(t)}
        -
        \lambda_d^{(t)}
    \right)s.
\]
Substituting $s=\|\sin\Theta_{\ell}^{(t)}\|_F^2$
gives Eq.~\ref{eq:app_staleness_angle_bound}.
\end{proof}

Therefore, degradation of a frozen dictionary is directly tied to movement of
the reconstruction-relevant subspace. A covariance shift that changes only
eigenvalues need not make the dictionary stale; a shift that rotates the
principal subspace does.

\subsection{Covariance Perturbations and Subspace Drift}
\label{app:covariance_perturbation}

Write the covariance at generation $t$ as
\[
    \Sigma_{\ell}^{(t)}
    =
    \Sigma^{(0)}
    +
    \Delta_{\ell}^{(t)}.
\]
Let
\[
    \gamma^{(0)}
    =
    \lambda_k(\Sigma^{(0)})
    -
    \lambda_{k+1}(\Sigma^{(0)})
\]
be the generation-$0$ eigengap.

Standard eigenspace perturbation results imply that, while the perturbation
remains small relative to this eigengap,
\begin{equation}
    \left\|
        \sin\Theta_{\ell}^{(t)}
    \right\|_F
    \lesssim
    \frac{
        \sqrt{k}
        \|\Delta_{\ell}^{(t)}\|_{\mathrm{op}}
    }{
        \gamma^{(0)}
    }.
    \label{eq:app_davis_kahan}
\end{equation}

Combining Eqs.~\ref{eq:app_staleness_angle_bound}
and~\ref{eq:app_davis_kahan} gives
\begin{equation}
    S_{\ell}(t)
    \lesssim
    \left(
        \lambda_1^{(t)}
        -
        \lambda_d^{(t)}
    \right)
    \frac{
        k\|\Delta_{\ell}^{(t)}\|_{\mathrm{op}}^2
    }{
        (\gamma^{(0)})^2
    }.
    \label{eq:app_staleness_covariance}
\end{equation}

The bound is not intended to predict the exact empirical magnitude of SAE
damage. Rather, it establishes the qualitative mechanism needed by the main
paper: cumulative covariance shift can move the optimal reconstruction
subspace, causing an SAE trained at generation $0$ to become increasingly
mismatched.

It also highlights the role of spectral structure. For the same covariance
perturbation, a representation with a small eigengap can undergo greater
subspace movement than one with a well-separated principal subspace.

\subsection{Matched-Control Decomposition}
\label{app:matched_decomposition}

The experiments compare every synthetic checkpoint against a real-data
checkpoint receiving the same continued-training budget. The same comparison
can be made in the linear model.

For reconstruction error, define
\begin{align}
    D_{\mathrm{zs}}(t)
    &=
    \mathcal{L}^{\mathrm{zs}}_{\mathrm{syn}}(t)
    -
    \mathcal{L}^{\mathrm{zs}}_{\mathrm{real}}(t),\\
    D_{\mathrm{adapt}}(t)
    &=
    \mathcal{L}^{\mathrm{adapt}}_{\mathrm{syn}}(t)
    -
    \mathcal{L}^{\mathrm{adapt}}_{\mathrm{real}}(t).
\end{align}

Since
\[
    \mathcal{L}^{\mathrm{zs}}_{\ell}(t)
    =
    \mathcal{L}^{\mathrm{adapt}}_{\ell}(t)
    +
    S_{\ell}(t),
\]
we obtain the exact decomposition
\begin{equation}
    D_{\mathrm{zs}}(t)
    -
    D_{\mathrm{adapt}}(t)
    =
    S_{\mathrm{syn}}(t)
    -
    S_{\mathrm{real}}(t).
    \label{eq:app_damage_decomposition}
\end{equation}

Equation~\ref{eq:app_damage_decomposition} gives a direct interpretation of
the gap between zero-shot and adapted evaluation. If self-consumption causes
more dictionary mismatch than matched continued training on real data, then
\[
    S_{\mathrm{syn}}(t)
    >
    S_{\mathrm{real}}(t),
\]
and adapting the dictionary reduces the synthetic--real reconstruction gap.

Importantly, a small value of $D_{\mathrm{adapt}}(t)$ does not imply that
$\Sigma_{\mathrm{syn}}^{(t)}$ is unchanged. It only means that, after allowing
the reconstruction subspace to adapt, the synthetic representation is not
substantially harder to reconstruct than the matched real-data representation.

\subsection{Reconstruction Does Not Identify Individual Features}
\label{app:rotation}

The preceding analysis concerns the subspace captured by an autoencoder.
Feature-level evaluations, however, depend on how information is arranged
\emph{within} that subspace.

This distinction can be seen even in the linear setting. Let
$W\in\mathbb{R}^{d\times k}$ be an orthonormal decoder and let
$Q\in\mathbb{R}^{k\times k}$ be any orthogonal matrix. Define
\[
    W' = WQ.
\]
Then
\begin{equation}
    W'W'^\top
    =
    WQQ^\top W^\top
    =
    WW^\top.
    \label{eq:app_rotation_reconstruction}
\end{equation}
Thus $W$ and $W'$ give exactly the same reconstruction for every input.

Their latent coordinates are nevertheless different:
\[
    z=W^\top h,
    \qquad
    z'=W'^\top h=Q^\top z.
\]

\begin{proposition}[Full-space probing is invariant to rotations]
\label{prop:probe_invariance}
For any linear predictor $a^\top z$ defined on the original latent
representation, there exists a predictor on $z'$ that gives exactly the same
output for every input.
\end{proposition}

\begin{proof}
Choose
\[
    a' = Q^\top a.
\]
Then
\[
    (a')^\top z'
    =
    a^\top QQ^\top z
    =
    a^\top z.
\]
\end{proof}

Thus a probe with access to the complete latent representation can recover the
same information after an invertible orthogonal change of coordinates.

Sparse coordinate-based probes do not have this invariance. Consider, for
example, a two-dimensional representation where the task-relevant signal is
localized entirely in $z_1$. Under the rotation
\[
    Q
    =
    \frac{1}{\sqrt{2}}
    \begin{bmatrix}
        1 & -1\\
        1 &  1
    \end{bmatrix},
\]
the same signal is distributed across both rotated coordinates. Reconstruction
is unchanged, and a probe using both coordinates loses no information, but
neither individual coordinate need retain the original level of predictive
power.

This example establishes a one-way implication:
\[
    \text{good reconstruction}
    \centernot\Rightarrow
    \text{information localized in individual latents}.
\]
It does \emph{not} imply that SAE fine-tuning necessarily rotates features or
necessarily reduces sparse localization. Rather, it shows why reconstruction
recovery alone cannot establish recovery of feature-level usefulness.

\subsection{An Illustrative Recursive Covariance Model}
\label{app:recursive_model}

We next revisit a simple recursive Gaussian model to illustrate how repeated
training can accumulate covariance perturbations. This model is deliberately
stylized and is not intended as a mechanistic description of language-model
self-consumption.

Let the generation-$0$ distribution be
\[
    h^{(0)}
    \sim
    \mathcal{N}(0,\Sigma^{(0)}).
\]

Consider first the idealized population update
\begin{equation}
    \Sigma_\star^{(t+1)}
    =
    \Sigma_\star^{(t)}
    +
    \sigma^2 I,
    \qquad
    \Sigma_\star^{(0)}
    =
    \Sigma^{(0)}.
    \label{eq:app_isotropic_loop}
\end{equation}
Hence
\begin{equation}
    \Sigma_\star^{(t)}
    =
    \Sigma^{(0)}
    +
    t\sigma^2 I.
    \label{eq:app_isotropic_solution}
\end{equation}

\subsubsection{Isotropic population drift is not sufficient for dictionary staleness}

If
\[
    \Sigma^{(0)}
    =
    V\Lambda^{(0)}V^\top,
\]
then
\[
    \Sigma_\star^{(t)}
    =
    V
    \left(
        \Lambda^{(0)}+t\sigma^2 I
    \right)
    V^\top.
\]
The eigenvectors are therefore identical at every generation, and
\[
    P_\star^{(t)}=P^{(0)}.
\]

Consequently,
\[
    S_\star(t)=0.
\]

This observation is useful because it identifies what the theory requires:
\emph{variance growth alone does not make a linear dictionary stale}. The
staleness observed in the experiments requires a shift that changes the
reconstruction-relevant geometry, such as anisotropic covariance drift,
finite-sample perturbations, or other changes induced by neural-network
training.

\subsubsection{Finite-sample perturbations}

Suppose that at each generation only $n$ samples are available. Let
$\widehat{\Sigma}^{(t)}$ denote the empirical covariance and define
\[
    E^{(t)}
    =
    \widehat{\Sigma}^{(t)}
    -
    \Sigma_\star^{(t)}.
\]
The actual covariance used to generate the next sample is
\[
    \widetilde{\Sigma}^{(t)}
    =
    \widehat{\Sigma}^{(t-1)}
    +
    \sigma^2 I.
\]

Conditional on $\widetilde{\Sigma}^{(t)}$, standard concentration for Gaussian
sample covariance gives, with high probability,
\begin{equation}
    \left\|
        \widehat{\Sigma}^{(t)}
        -
        \widetilde{\Sigma}^{(t)}
    \right\|_{\mathrm{op}}
    \leq
    \epsilon_{n,d,\delta}
    \left\|
        \widetilde{\Sigma}^{(t)}
    \right\|_{\mathrm{op}},
    \label{eq:app_single_step}
\end{equation}
where
\[
    \epsilon_{n,d,\delta}
    =
    C\left[
        \sqrt{
            \frac{d+\log(1/\delta)}{n}
        }
        +
        \frac{d+\log(1/\delta)}{n}
    \right]
\]
for an absolute constant $C$.

Because
\[
    \widetilde{\Sigma}^{(t)}
    -
    \Sigma_\star^{(t)}
    =
    E^{(t-1)},
\]
we have
\begin{align}
    \|E^{(t)}\|_{\mathrm{op}}
    &\leq
    \left\|
        \widehat{\Sigma}^{(t)}
        -
        \widetilde{\Sigma}^{(t)}
    \right\|_{\mathrm{op}}
    +
    \|E^{(t-1)}\|_{\mathrm{op}}\\
    &\leq
    (1+\epsilon)
    \|E^{(t-1)}\|_{\mathrm{op}}
    +
    \epsilon
    \left(
        \lambda_1^{(0)}
        +
        t\sigma^2
    \right),
    \label{eq:app_recursive_error}
\end{align}
where $\epsilon=\epsilon_{n,d,\delta/T}$ after a union bound over
generations.

If $E^{(0)}=0$, unrolling the recursion gives
\begin{equation}
    \|E^{(t)}\|_{\mathrm{op}}
    \leq
    \epsilon
    \sum_{s=1}^{t}
    (1+\epsilon)^{t-s}
    \left(
        \lambda_1^{(0)}
        +
        s\sigma^2
    \right).
    \label{eq:app_accumulated_error}
\end{equation}

Unlike the isotropic population term, $E^{(t)}$ is generally anisotropic and
can therefore rotate empirical principal directions. Applying
Eq.~\ref{eq:app_davis_kahan} to this perturbation shows how recursive
finite-sample error can create an increasingly stale empirical dictionary.

We emphasize that this construction is an illustrative example of the more
general covariance-drift mechanism in Sec.~\ref{sec:theory}. In the language
models studied experimentally, representational changes can arise from many
sources beyond finite-sample covariance estimation.

\subsection{Threshold-Based Sparsity Under Variance Growth}
\label{app:sparsity_toy}

The isotropic population model can nevertheless illustrate how activation
scale alone affects a threshold-based notion of sparsity.

For the $i$th principal component,
\[
    z_i^{(t)}
    =
    v_i^\top h^{(t)},
\]
and under Eq.~\ref{eq:app_isotropic_solution},
\[
    z_i^{(t)}
    \sim
    \mathcal{N}
    \left(
        0,
        \lambda_i^{(0)}+t\sigma^2
    \right).
\]

For a fixed firing threshold $\tau>0$,
\begin{align}
    p_i^{(t)}
    &:=
    \Pr\left(
        |z_i^{(t)}|>\tau
    \right)\\
    &=
    2\Phi
    \left(
        -
        \frac{\tau}{
            \sqrt{
                \lambda_i^{(0)}+t\sigma^2
            }
        }
    \right),
    \label{eq:app_firing_probability}
\end{align}
where $\Phi$ is the standard normal CDF.

Since the variance increases with $t$, $p_i^{(t)}$ increases monotonically.
Equivalently, the fraction of inactive coordinates
\begin{equation}
    \Psi^{(t)}
    =
    1-
    \frac{1}{k}
    \sum_{i=1}^{k}
    p_i^{(t)}
    \label{eq:app_sparsity_index}
\end{equation}
decreases with generation. This calculation provides a simple mechanism by which increasing activation variance can increase a threshold-based $L_0$ quantity even without a rotation of the underlying subspace. It should therefore be interpreted separately from the staleness analysis above. In particular, because isotropic variance growth leaves the optimal linear dictionary unchanged, this model alone cannot explain why SAE fine-tuning reduces the $L_0$ gap in the experiments.

\subsection{Scope of the Theory}
\label{app:theory_scope}

The analysis above is intentionally limited. First, a linear autoencoder is a proxy for an SAE, not an exact model of a nonlinear sparse dictionary. The PCA equivalence is used to obtain a tractable notion of reconstruction-relevant representational drift. Second, the theory does not predict that self-consumption must produce monotonic degradation in every SAE metric. It predicts that movement of the reconstruction-relevant subspace creates a penalty for a frozen dictionary and that adapting the dictionary can remove this penalty. Third, reconstruction recovery places no guarantee on the organization of individual features. Consequently, the theory is compatible both with the persistent sparse-localization degradation observed in Pythia and with the more heterogeneous feature-level effects observed across Gemma layers. Finally, the theory does not explain why particular model layers drift more than others. The depth dependence observed empirically should therefore be viewed as an experimental finding: the analysis states that greater subspace drift produces greater dictionary staleness, but does not predict which layers should undergo that drift.

\end{document}